\pdfoutput=1
\documentclass[11pt]{article}
\usepackage{acl}
\usepackage{times}
\usepackage{latexsym}
\usepackage[T1]{fontenc}
\usepackage[utf8]{inputenc}
\usepackage{microtype}
\usepackage{inconsolata}
\usepackage{graphicx}
\usepackage{adjustbox}

\title{L3iTC at the FinLLM Challenge Task: Quantization for Financial Text Classification \& Summarization}

\author{Elvys Linhares Pontes$^{1}$ \and Carlos-Emiliano González-Gallardo$^{2}$\\ \and \textbf{Mohamed Benjannet$^{1}$} \and \textbf{Caryn Qu$^{1}$} \and \textbf{Antoine Doucet$^{2}$} \\
         {\bf$^{1}$} Trading Central Labs, Trading Central, Paris, France \\ {\bf$^{2}$} University of La Rochelle, L3i, La Rochelle, France \\ 
         \small{\texttt{\{elvys.linhares\_pontes,mohamed.benjannet,caryn.qu\}@tradingcentral.com}}\\
         \small{\texttt{\{carlos.gonzalez\_gallardo,antoine.doucet\}@univ-lr.fr}}
}

\begin{document}
\maketitle
\begin{abstract}
This article details our participation (L3iTC) in the FinLLM Challenge Task 2024, focusing on two key areas: Task 1, financial text classification, and Task 2, financial text summarization. To address these challenges, we fine-tuned several large language models (LLMs) to optimize performance for each task. Specifically, we used 4-bit quantization and LoRA to determine which layers of the LLMs should be trained at a lower precision. This approach not only accelerated the fine-tuning process on the training data provided by the organizers but also enabled us to run the models on low GPU memory. Our fine-tuned models achieved third place for the financial classification task with an F1-score of $0.7543$ and secured sixth place in the financial summarization task on the official test datasets.
\end{abstract}

\section{Introduction}

Financial markets are characterized by their complexity and the vast volume of unstructured data they generate daily. The use of Large Language Models (LLMs) in finance has brought significant focus to tasks involving the analysis, generation, and decision-making related to financial texts. Indeed, LLMs have demonstrated remarkable performance in a large range of applications, from conversational agents to complex decision-making systems.
Despite the advances, their potential for thorough analysis and decision-making in finance is still unexplored.

The Financial Challenges in Large Language Models (FinLLM)\footnote{\url{https://sites.google.com/nlg.csie.ntu.edu.tw/finnlp-agentscen/shared-task-finllm}} aims to investigate and enhance the role of LLMs in advancing financial analysis and decision-making processes~\cite{xie2024finben}. More precisely, it focuses on three applications: financial classification of sentences~\cite{sy-etal-2023-fine}, financial news summarization~\cite{zhou2021trade}, and single stock trading~\cite{yu2023finmem}.

Motivated by these challenges, we participated (L3iTC) in the financial text classification and financial text summarization tasks. We proposed a fine-tuning process that combines 4-bit quantization and LoRA to optimize several LLMs for each task. This approach accelerated the fine-tuning process on the training data provided by the organizers but also enabled us to run the models on low GPU memory. Our results secured the third place for the financial classification task with an F1-score of $0.7543$ and sixth place in financial text summarization on the official test datasets.


\section{FinLLM Challenge Task}

With the advent of LLMs in finance, tasks related to financial text analysis, generation, and decision-making have garnered increasing attention. Key applications in this domain include financial classification, financial text summarization, and single-stock trading. While several approaches utilizing LLMs have demonstrated remarkable performance in these areas, their capabilities for comprehensive analysis and decision-making in finance remain largely unexplored.

FinLLM aims to investigate and enhance the role of LLMs in advancing financial analysis and decision-making processes~\cite{xie2024finben}.

\subsection{Task 1: Financial Classification}

The first task aims to evaluate the capabilities of LLMs in identifying and categorizing texts as either premises or claims \cite{sy-etal-2023-fine}. This task is particularly challenging due to financial texts' nuanced and complex nature, where distinguishing between these concepts (claims and premises) requires sophisticated understanding and contextual analysis. The organizers provided a training dataset with 7.75k data examples and the official test dataset composed of 969 examples.

\subsection{Task 2: Financial Text Summarization}

This task is designed to test the capabilities of LLMs in generating coherent and concise summaries~\cite{zhou2021trade}. The challenge lies in the ability to accurately capture the essential points and nuances of complex financial news, ensuring that the summary remains both informative and coherent~\cite{10214663}. The organizers provided a training dataset with 8k data examples and the official test dataset composed of 2k examples.





\subsection{Model Leakage Detection}

To measure the risk of data leakage from the test set during model training, organizers have developed a new metric called the Data Leakage Test (DLT), building on existing research~\cite{wei2023skywork}. DLT assesses the risk of data leakage by calculating the difference in perplexity between training and test data for large language models (LLMs). A larger DLT value indicates a lower likelihood of the LLM having seen the test set during training, suggesting a lower risk of model cheating, while a smaller DLT value suggests a higher risk of data leakage and model cheating.

\subsection{Evaluation Metrics}

For the financial text classification task, the organizers employed two primary evaluation metrics to gauge the performance of the participants' models: F1-score and accuracy. F1-score (F1) considers both precision and recall, providing a balanced measure of a model's accuracy. Accuracy represents the ratio of correctly predicted instances to the total instances.

For the financial text summarization task, organizers used ROUGE (1, 2, and L), BERTScore, and BARTScore metrics. ROUGE-n measures the overlap of n-grams between the generated summaries and the reference summaries. BERTScore calculates the similarity between the generated and reference summaries using sentence representation. Finally, BARTScore compares the generated summaries against a reference summary to determine how well the generated summaries capture the reference summaries' meaning, fluency, and coherence.

\section{L3iTC Approaches}

\begin{figure}[h]
\centering
  \includegraphics[width=0.8\columnwidth]{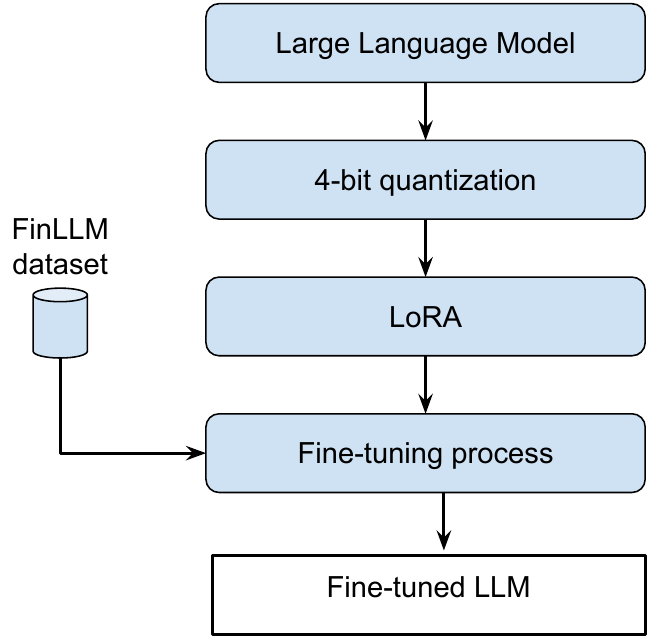}
  \caption{Fine-tuning methodology}
  \label{fig:fine-tune}
\end{figure}

We participated in the first two tasks. We developed the following architecture to address these tasks to generate our fine-tuned LLM for the FinLLM shared task (Figure~\ref{fig:fine-tune}).

\subsection{Large Language Models}

LLMs can perform a variety of natural language processing tasks such as translation, summarization, and conversational dialogue~\cite{chang2024survey}. They are trained on diverse datasets encompassing a wide range of topics, enabling them to generate coherent and contextually relevant responses. Among prominent LLMs available today, we selected the following Instruct models due to their high performance and relative small size: \textrm{Mistral-7B-Instruct-v0.2}\footnote{\url{https://huggingface.co/mistralai/Mistral-7B-Instruct-v0.2}}, \textrm{Mistral-7B-Instruct-v0.3}\footnote{\url{https://huggingface.co/mistralai/Mistral-7B-Instruct-v0.3}}, and \textrm{Meta-Llama-3-8B-Instruct}\footnote{\url{https://huggingface.co/meta-llama/Meta-Llama-3-8B-Instruct}}.

\subsection{Fine-tuning}

In the classic fine-tuning of LLMs, a significant portion of model weights is typically modified, necessitating substantial computational resources. To alleviate GPU memory requirements during fine-tuning, we employed quantization techniques as proposed by \citet{NEURIPS2022_c3ba4962}. Specifically, we utilized 4-bit quantization to reduce the memory footprint of LLMs prior to fine-tuning. 

To make fine-tuning more efficient, LoRA~\cite{hu2021lora} improves efficiency by using two smaller matrices, known as update matrices, to represent weight updates via low-rank decomposition. These matrices are trained to adjust to new data while minimizing the total number of changes. The original weight matrix stays unchanged and is not further modified. The final results are achieved by combining the original weights with the adapted ones. In our study, we focused on training parameters within specific modules, including ``q\_proj'', ``k\_proj'', ``v\_proj'', ``o\_proj'', ``gate\_proj'', ``up\_proj'', ``down\_proj'', and ``lm\_head'' employing a dropout rate of 0.05.

For both tasks undertaken, we partitioned the training dataset into three subsets: train, validation, and test. The validation and test subsets each comprised 10\% of the examples, with the remaining 80\% constituting the training data. We set the learning rate to $5 \times 10^{-5}$ and the batch size to 4. The models underwent fine-tuning over 2,000 steps.

\section{Preliminary results}

Table~\ref{tb:task1-finetune} summarizes the performance of fine-tuning each LLM for Task 1. Initially, the models predicted more than just the target word, attempting to choose the correct class and justify their selection. This approach led to poor results. The best model without fine-tuning was \textrm{Mistral-7B-Inst-v0.2}, which achieved an accuracy of $54\%$ and an F1-score of $0.39$.

Despite the limitations encountered during the fine-tuning process, particularly those related to LoRA configuration and 4-bit quantization, all fine-tuned models showed improved performance by generating only the target predicted class. Notably, the best fine-tuned model was \textrm{FT-Clas-Mistral-7B-Inst-v0.3}, which achieved an accuracy of $78\%$ and an F1-score of $0.78$. Therefore, we selected it to compete on the first task on the official test dataset.

\begin{table}[h]
\begin{adjustbox}{width=\linewidth}
\begin{tabular}{lcc}
\hline
\textbf{Team}           & \textbf{Accuracy} & \textbf{F1} \\
\hline
Mistral-7B-Inst-v0.2  & 54\% & 0.39 \\
Mistral-7B-Inst-v0.3  & 46\% & 0.36 \\
Meta-Llama-3-8B-Inst  & 52\% & 0.48 \\
\hline
FT-Clas-Mistral-7B-Inst-v0.2  & 76\% & 0.76 \\
\textbf{FT-Clas-Mistral-7B-Inst-v0.3}  & \textbf{78\%} & \textbf{0.78} \\
FT-Clas-Meta-Llama-3-8B-Inst  & 67\% & 0.67 \\
\hline
\end{tabular}
\end{adjustbox}
\caption{Preliminary fine-tuning results for financial classification task. The selected model is in bold.}
\label{tb:task1-finetune}
\end{table}

Table~\ref{tb:task2-finetune} summarizes the performance of fine-tuning each LLM for Task 2. Unfortunately, the fine-tuning process did not yield significant improvements in the ROUGE score and even resulted in a decline in BERTScore performance. The main reasons for the poor results are mainly related to our finetuning process. More precisely, the quantization process of 4-bits indeed reduces the amount of GPU memory necessary to fine-tune the model; however, this quantization limited the precision of the learning process which also affected the quality of our models.

Although \textrm{FT-Sum-Mistral-7B-Inst-v0.2} obtained the best ROUGE-1 score, which is used as the final ranking metric, we found that BertScore better correlates summary quality and human judgment (Table~\ref{tb:app} lists some summaries generated by \textrm{FT-Sum-Mistral-7B-Inst-v0.2} and \textrm{Mistral-7B-Inst-v0.3}). Thus, we selected the \textrm{Mistral-7B-Inst-v0.3} model for the second task.

\begin{table}[h]
\centering
\begin{adjustbox}{width=\linewidth}
\begin{tabular}{lcc}
\hline
\textbf{Team}            & \textbf{ROUGE-1} & \textbf{BertScore}\\
\hline
Mistral-7B-Inst-v0.2     & 0.2245 & 0.5373 \\ 
\textbf{Mistral-7B-Inst-v0.3} & \textbf{0.2248} & \textbf{0.5374} \\ 
Meta-Llama-3-8B-Inst     & 0.2240 & 0.5333 \\ 
\hline
FT-Sum-Mistral-7B-Inst-v0.2  & 0.2312 & 0.5097 \\ 
FT-Sum-Mistral-7B-Inst-v0.3  & 0.2250 & 0.502 \\
FT-Sum-Meta-Llama-3-8B-Inst  & 0.2231 & 0.488 \\
\hline
\end{tabular}
\end{adjustbox}
\caption{Preliminary fine-tuning results for financial text summarization task. The selected model is highlighted in bold.}
\label{tb:task2-finetune}
\end{table}

\section{Official Results}

The organizers created a test dataset consisting of 969 test cases for the first task and 2,000 test cases for the second task.
The official results are listed in Tables~\ref{tb:task1-official} and~\ref{tb:task2-official} for tasks 1 and 2, respectively. For the financial classification task, our fine-tuned model achieved notable results, with an accuracy of $75.44\%$ and an F1-score of $0.7543$. Remarkably, we secured third place, trailing the first place by 0.0069 points in F1-score.

\begin{table}[h]
\begin{adjustbox}{width=\linewidth}
\begin{tabular}{lcccc}
\hline
\textbf{Team}           & \textbf{Accuracy} & \textbf{MCC}     & \textbf{F1} & \textbf{DLT}   \\
\hline
\textbf{Team Barclays}  & \textbf{76.26}\%   & \textbf{0.5237} & \textbf{0.7612} & \textbf{38.9} \\
Albatross      & 75.75\%   & 0.5174 & 0.7575 & -- \\
\textit{L3iTC} & \textit{75.44}\%   & \textit{0.5149} & \textit{0.7543} & \textit{2.2} \\
Wealth Guide   & 75.13\%   & 0.5018 & 0.7509 & -- \\
Finance Wizard & 72.86\%   & 0.4554 & 0.7262 & -- \\
CatMemo        & 71.10\%   & 0.4199 & 0.7086 & -- \\
Upaya          & 70.90\%   & 0.4166 & 0.7083 & -- \\

\hline
\end{tabular}
\end{adjustbox}
\caption{Official results for the Financial classification task on the test dataset (Task 1). The leaderboard is ranked by F1-score (F1) scores, with the top team highlighted in bold and our team highlighted in italics.}
\label{tb:task1-official}
\end{table}

\begin{table*}[h!]
\centering
\begin{tabular}{lcccccc}
\hline
\textbf{Team}           & \textbf{ROUGE-1} & \textbf{ROUGE-2} & \textbf{ROUGE-L} & \textbf{BertScore} & \textbf{BartScore} & \textbf{DLT}\\
\hline
\textbf{LBZ}   & \textbf{0.5346}  & \textbf{0.3581}  & \textbf{0.4921}  & \textbf{0.9117}    & \textbf{-3.41} \\
Upaya          & 0.5294  & 0.3582  & 0.4860   & 0.9106    & -3.45 & -- \\
Finance Wizard & 0.5210   & 0.3406  & 0.4735  & 0.9083    & -3.49 & 0.8332\\
Revelata       & 0.5004  & 0.3330   & 0.4643  & 0.9070     & -3.80 & 1.7346 \\
Albatross      & 0.3691  & 0.2010   & 0.3227  & 0.8720     & -3.93 & -- \\
\textit{L3iTC} & \textit{0.3661}  & \textit{0.1872}  & \textit{0.3046}  & \textit{0.8750}     & \textit{-4.25} & -- \\
Wealth Guide   & 0.3089  & 0.1794  & 0.2819  & 0.8595    & -4.96 & -- \\
\hline
\end{tabular}
\caption{Official results for the Financial text summarization task on the test dataset (Task 2). The leaderboard is ranked by ROUGE-1 scores, with the top team highlighted in bold and our team highlighted in italics.}
\label{tb:task2-official}
\end{table*}

\begin{table*}[h!]
\centering
\begin{adjustbox}{width=0.93\linewidth}
\begin{tabular}{p{.1\textwidth}p{.35\textwidth}p{.5\textwidth}}
\hline
\textbf{Id}           & \textbf{Mistral-7B-Inst-v0.3} & \textbf{FT-Sum-Mistral-7B-Inst-v0.2} \\
\hline
\textbf{edtsum0}   & The all-season tire market in Europe is projected to grow by USD 3.42 billion during 2020-2024, with a CAGR of nearly 9\% ...  & All-Season Tire Market in Europe 2020-2024 | Evolving Opportunities with Bridgestone Corp. and Continental AG | Technavio ... \\

\textbf{edtsum2}   & The chocolate market is projected to grow by USD 41.15 billion from 2020 to 2024, with a CAGR of over 5\% ...  & Chocolate Market to Grow by USD 41.15 Billion During 2020-2024 | Increasing Premiumization of Chocolates to boost the Market Growth | Technavio Report | English USA - English USA - English USA - ... \\

\textbf{edtsum15}   & Aon PLC has experienced dealings with Farallon Capital Management, L.L.C. on behalf of funds managed jointly with Farallon Partners, L.L.C. ...  & Form 8.3 - Aon PLC - 12 April 2021 - Farallon Capital Management, L.L.C. on behalf of funds managed jointly with Farallon Partners, L.L.C. - Amended - 13 April 2021 - 10:00 am - 10:00 am  \\

\hline
\end{tabular}
\end{adjustbox}
\caption{Examples of the generated summaries on the official test dataset. These examples demonstrate that the readability and informativeness are superior in the \textrm{Mistral-7B-Inst-v0.3} model compared to the \textrm{FT-Sum-Mistral-7B-Inst-v0.2} model.}
\label{tb:app}
\end{table*}

As expected, our models did not achieve the best results, primarily due to limitations encountered during the fine-tuning process. Specifically, the use of 4-bit quantization constrained the learning process and adversely affected the quality of summary generation. Consequently, our model ranked sixth for the second task, with a ROUGE-1 score of $0.3661$ and a BERTScore of $0.875$.

\section{Conclusion}

This article presents our participation (L3iTC) in the FinLLM Challenge Task 2024, concentrating on two primary tasks: Task 1, financial text classification, and Task 2, financial text summarization. To tackle these challenges, we fine-tuned several LLMs to enhance their performance for each task.

For Task 1, our fine-tuning efforts led to a third-place finish, achieving an F1-score of $0.7543$, just 0.0069 points behind the first place. In Task 2, we secured sixth place on the official test datasets. These outcomes demonstrate that combining LoRA configuration and 4-bit quantization allows for the efficient fine-tuning of LLMs minimizing GPU memory and processing time, yielding notable results for tasks that do not require the generation of numerous tokens. In addition, combining quantization and LoRA enables the possibility of fine-tuning LLMs in smaller infrastructures that demand less energy thus reducing their carbon footprint \cite{samsi2023words}. 
However, when the number of tokens generated increases, as in the case of financial text summarization, this approach reveals its limitations. The quality of the generated summaries declines compared to those produced by the original LLMs without fine-tuning, highlighting the trade-offs involved in using this combination for tasks requiring extensive text generation.

Future work will focus on enhancing the fine-tuning process by employing 8-bit or 16-bit quantization and evaluating their performance on complex tasks such as summarization. We also plan to pre-train our model using a variety of summarization datasets and incorporate datasets from diverse tasks. This approach aims to develop a more robust model capable of handling various tasks without compromising the quality of generation.

\section*{Acknowledgments}

This work has been supported by the ANNA (2019-1R40226), TERMITRAD (2020-2019-8510010), Pypa (AAPR2021-2021-12263410), and Actuadata (AAPR2022-2021-17014610) projects funded by the Nouvelle-Aquitaine Region (France); as well as the France Relance (ANR-21-PRRD-0010-01) project funded by the French National Recherche Agency (ANR). We would like to also thank José G. Moreno for the insightful discussions.

\bibliography{references}

\end{document}